\documentclass[10pt,twocolumn,letterpaper]{article}

\usepackage{iccv}
\usepackage{times}
\usepackage{epsfig}
\usepackage{graphicx}
\usepackage{amsmath}
\usepackage{amssymb}
\usepackage{algorithm}
\usepackage{algorithmic}
\usepackage{subfigure}
\usepackage[bottom]{footmisc}
\graphicspath{{images/}}

\usepackage[breaklinks=true,bookmarks=false]{hyperref}

\iccvfinalcopy 

\def\iccvPaperID{****} 

\ificcvfinal\fi

\begin{document}
	
	\title{ Hard-Aware Deeply Cascaded Embedding}
	\author{Yuhui Yuan$^{1,3}$ \hspace{6mm} Kuiyuan Yang$^{2}$  \hspace{6mm}   Chao Zhang$^{1,4}$\thanks{Corresponding author : Chao Zhang.} \\	
		$^{1}$Key Laboratory of Machine Perception(MOE),Peking University\\
		$^{2}$DeepMotion  \hspace{10mm}  $^{3}$Microsoft Research \\		
		$^{4}$Cooperative Medianet Innovation Center, Shanghai Jiao Tong University\\
		{\tt\small yhyuan@pku.edu.cn, kuiyuanyang@deepmotion.ai, chzhang@cis.pku.edu.cn}
	}
	
	\maketitle
	\thispagestyle{empty}
	
	\begin{abstract}
		Riding on the waves of deep neural networks, deep metric learning has achieved promising results in various tasks by using triplet network or Siamese network. Though the basic goal of making images from the same category closer than the ones from different categories is intuitive, it is hard to optimize the objective directly due to the quadratic or cubic sample size. Hard example mining is widely used to solve the problem, which spends the expensive computation on a subset of samples that are considered hard. However, hard is defined relative to a specific model. Then complex models will treat most samples as easy ones and vice versa for simple models, both of which are not good for training. It is difficult to define a model with the just right complexity and choose hard examples adequately as different samples are of diverse hard levels. This motivates us to propose the novel framework named Hard-Aware Deeply Cascaded Embedding(HDC) to ensemble a set of models with different complexities in cascaded manner to mine hard examples at multiple levels. A sample is judged by a series of models with increasing complexities and only updates models that consider the sample as a hard case. The HDC is evaluated on CARS196, CUB-200-2011, Stanford Online Products, VehicleID and DeepFashion datasets, and outperforms state-of-the-art methods by a large margin.
	\end{abstract}
	
	\vspace{-0.2cm}
	\section{Introduction}
	\vspace{-0.2cm}
	\begin{figure}[h]
		\centering
		\includegraphics[scale=0.6]{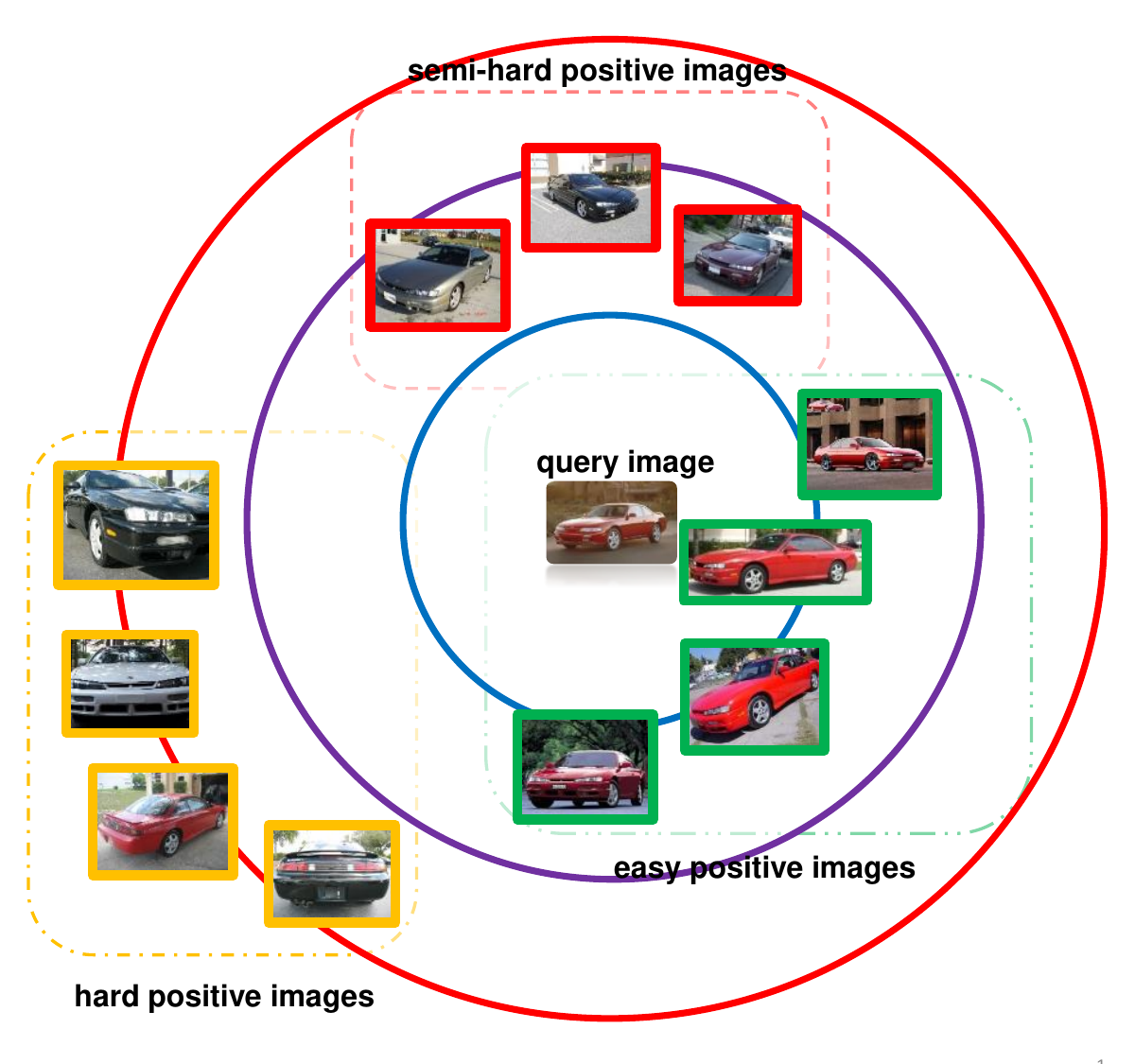}
		\caption{\textbf{Illustration of samples with different hard levels}: A query image is shown at the center, while other images from the same category (Nissan 240X Coupe 1998 from CARS196~\cite{stanford_cars}) are used to form positive pairs with the query image.}
		\label{fig:intro_car_distribution}
		\vspace{-0.5cm}
	\end{figure}
	
	Deep metric embedding has attracted increasing attention for various tasks , such as visual product search~\cite{cuhk_nips2016,DeepFashion,SimoSerraCVPR2016,song2015deep,UstinovaNIPS16}, face recognition~\cite{schroff2015facenet,tadmor2016learning,wen2016discriminative,bhattarai2016cp}, local image descriptor learning~\cite{han2015matchnet,softpn,G_2016_CVPR,simo2015discriminative}, person/vehicle re-identification~\cite{shi2016embedding,cheng2016person,you2016top,vehicleID}, zero-shot image classification~\cite{rippel2015metric,zhang2016zero,bucher2016improving}, fine-grained image classification~\cite{cui2015fine,wang2016mining,Zhang_2016_CVPR} and object tracking~\cite{leal2016learning,tao2016siamese}. Although deep metric embedding is modified into different forms for various tasks, it shares the same objective to learn an embedding space that pulls similar images closer and pushes dissimilar images far away. Typically, the target embedding space is learned with a convolutional neural network equipped with contrastive/triplet loss.
	
	Different from the traditional classification based models, the models of deep metric embedding consider two images (a pair) or three images (a triplet) as a training sample. Thus $N$ images can generate $\mathcal{O}(N^2)$ or $\mathcal{O}(N^3)$ samples. It becomes impossible to consider all samples even for a moderate number of images. Fortunately, not all samples are equally informative to train a model, which inspires many recent works to mine hard examples for training~\cite{cui2015fine,simo2015discriminative,wang2015unsupervised}.
	
	
	\begin{figure*}[htb]
		\centering
		\includegraphics[scale=0.45]{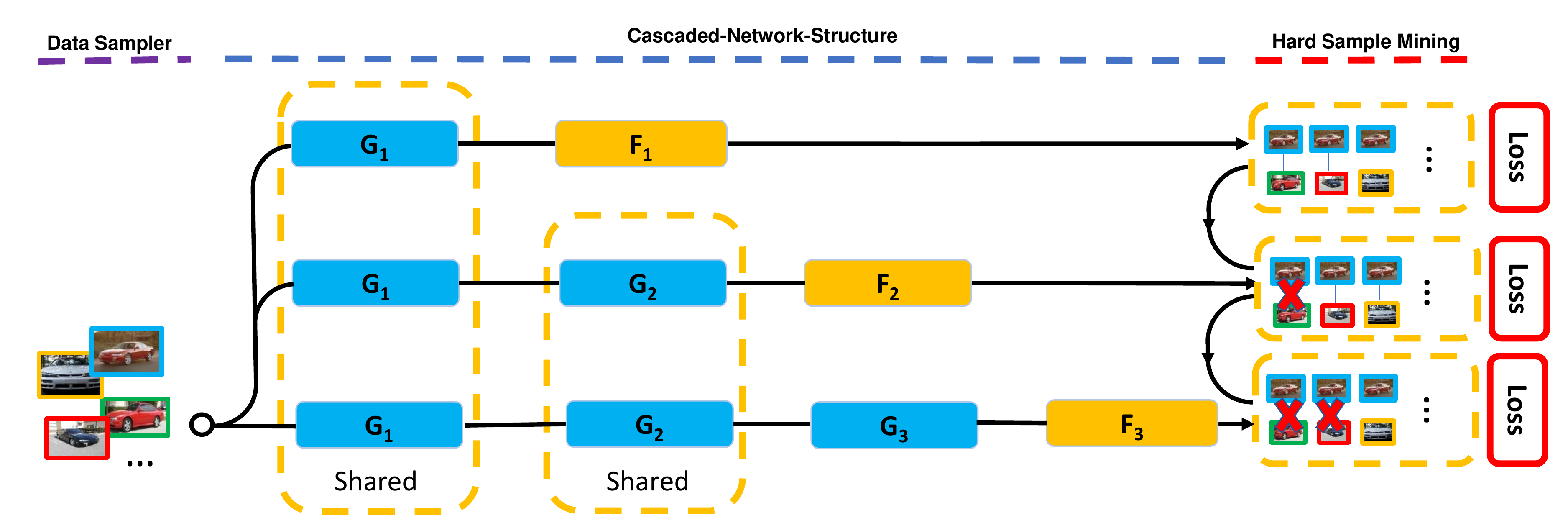}
		\caption{\textbf{Hard-Aware Deeply Cascaded Embedding} : We will train the first model with all the pairs, the second model with the semi-hard samples which are selected by the above model, the third model with the remained hard samples selected by the second model. Our framework support any K cascaded models. We plot the case=3 for convenience. $G_{1},G_{2},G_{3},F_{1},F_{2},F_{3}$ are the computation blocks in Convolutional Neural Networks.}
		\label{fig:our_method}
		\vspace{-0.35cm}
	\end{figure*}
	
	However, the hard level of a sample is defined relative to a model. Then samples can be divided into different hard levels as illustrated in Figure~\ref{fig:intro_car_distribution}. For a complex model, most samples will be treated as easy ones, and the model converges fast but is prone to overfitting. While for a simple model, most samples will be treated as hard ones and cannot fully benefit from hard example mining. It would be ideal to define a model with the just right complexity to mine hard examples adequately, which is an open problem itself.
	
	To alleviate the above problem, we ensemble a set of models with different complexities in a cascaded manner and mine hard examples adaptively, which is schematically illustrated in Figure~\ref{fig:our_method}. The most simple model is implemented by a shallow network, while complex models are implemented by cascading more layers following the simple ones. During the training phase, a sample will be considered by a series of models with increasing depth. Specifically, a sample firstly makes its forward pass through the simple model, the pass will stop if the simple model considers the sample as an easy one, otherwise the forward pass continues until a model considers the sample as an easy one or the deepest model is reached. Then the errors will be back-propagated to models that consider the sample as a hard case. We empirically show that the HDC achieves state-of-the-art results on five benchmarks.
	
	In summary, we make the following contributions:
	\begin{itemize}
		\vspace{-0.25cm}
		\item We propose the Hard-Aware Deeply Cascaded Embedding to solve the under-fitting and over-fitting problem when mining the hard samples during training. To the best of our knowledge, this is the first attempt to investigate and solve this problem.
		\vspace{-0.25cm}
		\item We conduct extensive experiments on five various datasets and all achieve state-of-the-art results. The promising results on different datasets demonstrate that the proposed method has good generalization capability.
	\end{itemize}
	
	\section{Related Work}
	\vspace{-0.2cm}
	Deep metric learning attracts great attention in recent years, and hard negative mining is becoming a common practice to effectively train deep metric networks~\cite{cui2015fine,simo2015discriminative,wang2015unsupervised}. Wang \etal~\cite{wang2015unsupervised} sample triplets during the first 10 training epoches randomly, and mine hard triplets in each mini-batch after 10 epoches. Cui \etal~\cite{cui2015fine} leverage human to label hard negative images from images assigned high confidence scores by the model during each round. Simo-Serra \etal~\cite{simo2015discriminative} analyze the influence of both of hard positive mining and hard negative mining, and find that the combination of aggressive mining for both positive and negative pairs improves the discrimination. However, these methods mine the hard images only based on a single model, which cannot adequately leverage samples with different hard levels.
	
	Our method of ensembling a set of models of different complexities in a cascaded manner shares the same spirit as the acceleration technique used in object/face detection~\cite{angelova2015real,huagang,pronet,yang2016exploit}. In the detection task, an image may contain several positive patches and a large number of negative patches. To reduce the computational cost, the model is broken down into a set of cascaded computation blocks, where computation blocks at early stages reject most easy background patches, while computation blocks at latter stages focus more on object-like patches.
	
	Our method also shares similar form with deeply-supervised network (DSN) proposed for image classification~\cite{Lee2014Deeply}, of which loss functions are added to the output layers and several middle layers. DSN improves the directness and transparency of the hidden layer learning process and tries to alleviate the ``gradient vanishing'' problem. Similar idea is adopted in GoogLeNet~\cite{Szegedy2015Going}. BranchyNet~\cite{teerapittayanon2016branchynet} attempts to speed up image classification by taking advantage of the DSN framework during test phase, where an image will be predicted using features learned at an early layer if high confidence score can be achieved. During the training phase of DSN, all samples are used and intermediate losses are only used to assist the training of the deepest model. While in our framework, samples of different hard levels are assigned to models with adequate complexities, and all models are ensembled together as a whole model. To be noted, ensemble is also a useful technique that has been widely used in model design to boost performance. Hinton \etal~\cite{srivastava2014dropout} add dropout into fully-connected layers, which implicitly ensembles an exponential number of sub-networks in a single network. He \etal~\cite{he2015deep} propose ResNet by adding residual connections into a network and win the \emph{ILSVRC 2015} competition, which is latterly proved by Veit \etal~\cite{veit2016residual} that ResNet is actually exponential ensembles of relatively shallow networks.
	
	In addition, there are several works focused on designing new loss functions for deep metric embedding recently. Rippel \etal~\cite{rippel2015metric} design a Nearest Class Multiple Centroids (NCMC) like loss which encourages images from the same category to form sub-clusters in the embedding space. Huang \etal~\cite{cuhk_nips2016} propose position-dependent deep metric to solve the problem that intra-class distance in a high-density region may be larger than the inter-class distance in low-density regions. Ustinova \etal~\cite{UstinovaNIPS16} propose a histogram loss, which aims to make the similarity distributions of positive and negative pairs less overlapping. Unlike other losses used for deep embedding, histogram loss comes with virtually no parameters that need to be tuned. K. Sohn \etal~\cite{sohn2016improved} proposed multi-class N-pair loss by generalizing triplet loss by allowing joint comparison among more than one negative examples. Different from our work, these works improve deep embedding by designing new loss functions within a single model. They can benefit from our method by mining hard examples adaptively using multiple cascaded models.
	
	\section{Hard-Aware Deeply Cascaded Embedding}
	\vspace{-0.2cm}
	Hard-Aware Deeply Cascaded embedding(HDC) is based on a straightforward intuition: handling samples of different hard levels with models of different complexities. Based on deep neural networks, models with different complexities can be naturally derived from sub-networks of different depths. For clarity, we will first formulate the general framework of HDC and then analyze the concrete case for the contrastive loss.
	
	\begin{figure*}[htb]
		\centering
		\subfigure[]{
			\label{fig:our_work_data:pos}
			{\includegraphics[scale=0.5]{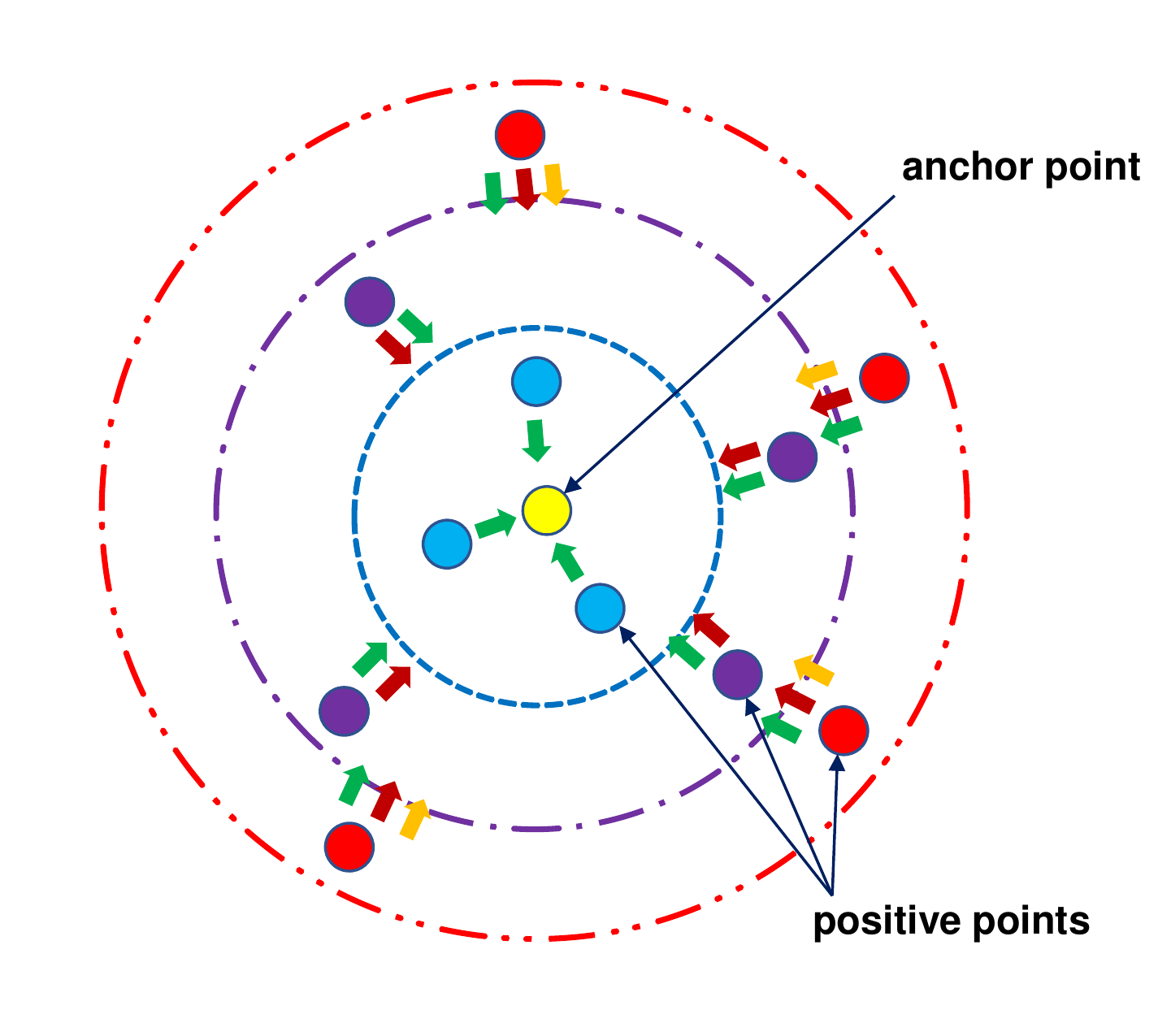}}}
		\hspace{12mm}
		\subfigure[]{
			\label{fig:our_work_data:neg}
			{\includegraphics[scale=0.49]{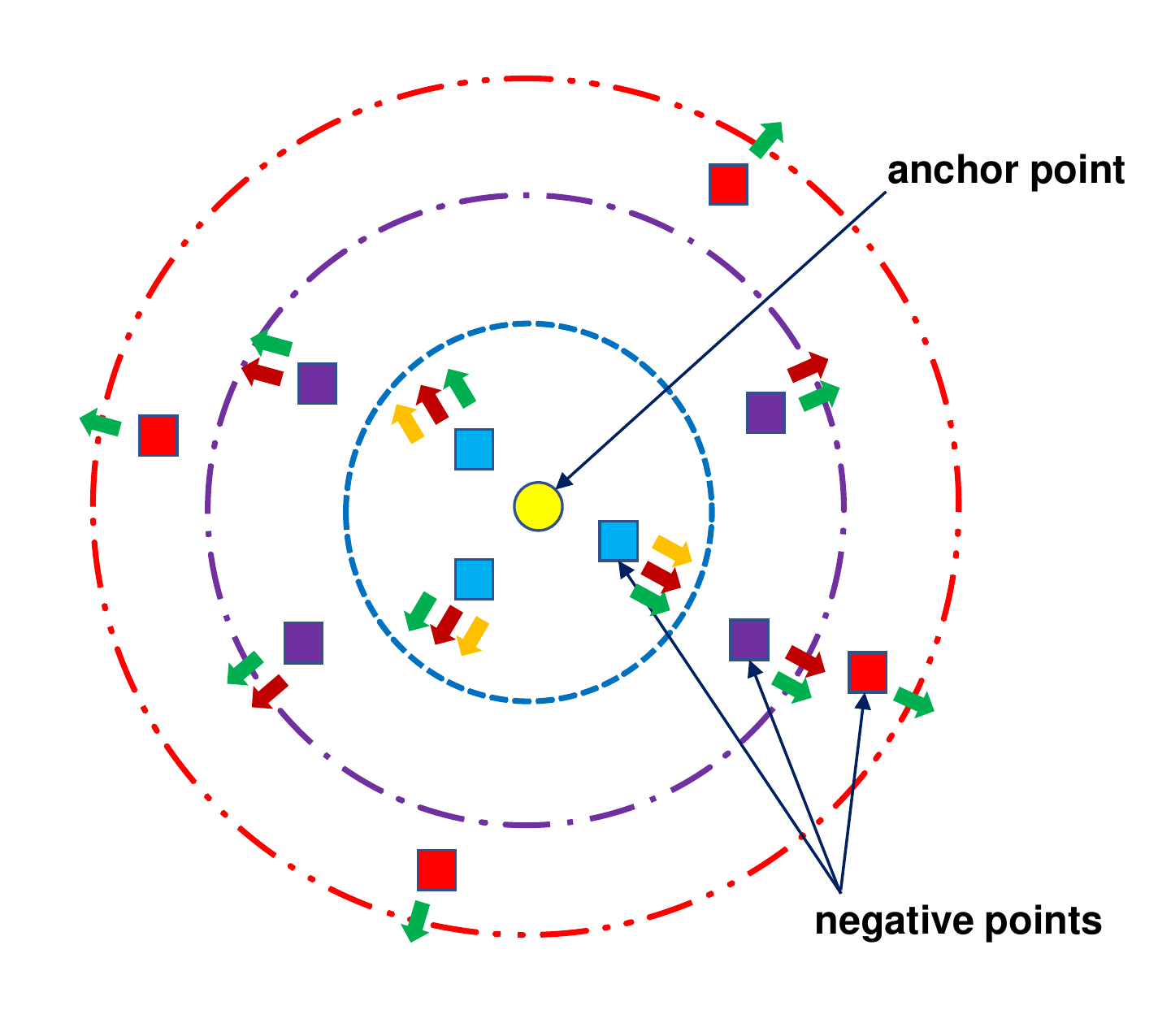}}}
		\label{fig:our_work_data}
		\caption{\textbf{Data Distribution:} (a) Positive Pairs Distribution: Based on the anchor point in the center, $\mathcal{P}_{0}$ contains all the points. $\mathcal{P}_{1}$ contains red, purple points. $\mathcal{P}_{2}$ only contains red points. (b) Negative Pairs Distribution: $\mathcal{N}_{0}$ contains all the points. $\mathcal{N}_{1}$ contains red, purple points. $\mathcal{N}_{2}$ only contains red points. Green arrows denote loss from Cascade-Model-1, red arrows denote loss from Cascade-Model-2 and yellow arrow denote loss from Cascade-Model-3.}
		\vspace{-0.35cm}
	\end{figure*}
	
	\subsection{ Model Formulation }
	\vspace{-0.2cm}
	Here are some notations that will be used to describe our method:
	\begin{itemize}
		\vspace{-0.25cm}
		\item \emph{$\mathcal{P}=\{I_i^{+}, I_j^{+}\}$} : all the positive image pairs constructed from training set, where $I_i^{+}$ and $I_j^{+}$ are supposed to be similar or share the same label.
		\vspace{-0.25cm}
		\item \emph{$\mathcal{N}=\{I_i^{-}, I_j^{-}\}$} : all the negative image pairs constructed from training set, where $I_i^{-}$ and $I_j^{-}$ are supposed to be irrelevant or from different labels.
		\vspace{-0.25cm}
		\item \emph{${G_k}$} : the $k^{th}$ computation block including several convolutional layers, pooling layers, and other possible operations in a network. Suppose there are $\rm{K}$ blocks in total, $G_1$ takes an image as input, and $G_k, k>1$ takes the outputs of its previous block as input, then all \rm{K} blocks are cascaded together as a feed-forward network.
		\vspace{-0.25cm}
		\item \emph{$\{o_{i,k}^{+}, o_{j,k}^{+}\}$} : the output of the $k^{th}$ computation block ${G_k}$ for the positive pairs $\{I_i^{+}, I_j^{+}\}$.
		\vspace{-0.25cm}
		\item \emph{$\{o_{i,k}^{-}, o_{j,k}^{-}\}$} : the output of the $k^{th}$ computation block ${G_k}$ for the negative pairs $\{I_i^{-}, I_j^{-}\}$.
		\vspace{-0.25cm}
		\item \emph{${F_k}$} : the $k^{th}$ transform function that transforms ${o_k}$ to a low dimensional feature vector ${f_k}$ for distance calculation.
		\vspace{-0.25cm}
		\item \emph{$\{f_{i,k}^{+}, f_{j,k}^{+}\}$} : the $k^{th}$ computed feature vector after ${F_k}$ for the positive pairs $\{I_i^{+}, I_j^{+}\}$.
		\vspace{-0.25cm}
		\item \emph{$\{f_{i,k}^{-}, f_{j,k}^{-}\}$} : the $k^{th}$ computed feature vector after ${F_k}$ for the negative pairs $\{I_i^{-}, I_j^{-}\}$.
	\end{itemize}
	\par
	Accordingly,  there are \rm{K} models corresponding to \rm{K} sub-networks of different depths. The first model is the simplest one which uses the first block $G_1$ and generates features for the pairs $\{I_i, I_j\}$  by:
	\begin{align}
	\{o_{i,1}, o_{j,1}\}&=G_{1}\circ\{I_i, I_j\}\\
	\{f_{i,1}, f_{j,1}\}&=F_{1}\circ\{o_{i,1}, o_{j,1}\}
	\end{align}
	\par
	If the pair is considered easy by the current model, it will not be passed to more complex models. Otherwise, the pair will continue its forward pass until the $k^{th}$ model considers it as an easy case or the final $\rm{K}^{th}$ model is reached. We can calculate the features of $k^{th}$ model by:
	\begin{align}
	\{o_{i,k}, o_{j,k}\}&=G_{k}\circ\{o_{i,k-1}, o_{j,k-1}\}\label{branch_feature_1}\\
	\{f_{i,k}, f_{j,k}\}&=F_{k}\circ\{o_{i,k}, o_{j,k}\}\label{branch_feature_3}
	\end{align}
	\par
	Then the loss of $k^{th}$ model is defined as:
	\begin{align}
	\mathcal{L}_{k} &= \sum_{(i,j) \in \mathcal{P}_{k}} \mathcal{L}_{k}^{+}(i,j) + \sum_{(i,j) \in \mathcal{N}_{k}} \mathcal{L}_{k}^{-}(i,j) \label{cascade_loss}
	\end{align}
	where $\mathcal{P}_{k}$ denotes all positive pairs that are considered as hard examples by previous models and $\mathcal{N}_k$ indicate negative ones. The definition of hard will be concretely given in Section~\ref{sec:hard_def}.
	
	Therefore, the final loss of the HDC is defined as:
	\begin{align}
	\mathcal{L} &= \sum_{k=1}^{\rm{K}}\lambda_{k}\mathcal{L}_{k}\label{cascade_sum_loss}
	\end{align}
	where $\lambda_k$ is the weight for model $k$.
	
	The HDC is different from previous deep metric embedding, where only a single model (i.e., model $\rm{K}$) is used to mine hard samples. As samples of a dataset are with diverse hard levels, it is difficult to find a single model with the just right complexity to mine hard samples. In contrast, the HDC framework cascades multiple models with increasing complexities and mines samples of different hard levels in a seamless way.
	
	
	The model parameters are distributed in $G_k$, $F_k$, $1\leq k \leq \rm{K}$. They can be optimized by the standard SGD, the gradient of $G_k$ is:
	\begin{align}
	\frac{\partial \mathcal{L}}{\partial G_{k}} = \sum_{l=k}^{\rm{K}} \lambda_{l} \frac{\partial \mathcal{L}_{l}}{\partial G_{k}}\label{cascade_bp_G}
	\end{align}
	where the gradient of $G_k$ is calculated by all the models that are built on $G_k$. The gradient of $F_k$ is:
	\begin{align}
	\frac{\partial \mathcal{L}}{\partial F_{k}} = \lambda_{k} \frac{\partial \mathcal{L}_{k}}{\partial F_{k}}\label{cascade_bp_F}
	\end{align}
	where the gradient of $F_k$ is only calculated by model $k$ since $F_k$ is only used by model $k$ for feature transformation.
	
	The HDC is general for deep metric embedding with hard example mining. Here we take contrastive loss as an example to give the specific loss function. We first introduce the original contrastive loss which penalizes large distance between positive pairs and negative pairs with distance smaller than a margin, i.e.,
	\begin{align}
	\mathcal{L}^{+}(i,j)&=\mathcal{D}({f}_{i}^{+}, {f}_{j}^{+}) \label{loss_pos_pair}\\
	\mathcal{L}^{-}(i,j)&=\max \{ 0, \mathcal{M} - \mathcal{D}({f}_{i}^{-}, {f}_{j}^{-})\label{loss_neg_pair}\}
	\end{align}
	where $\mathcal{D}(f_i, f_j)$ is the Euclidean distance between the two L2-normalized feature vectors of $f_i$ and $f_j$, $\mathcal{M}$ is the margin.
	By applying the contrastive loss to Eq.\eqref{cascade_loss}, we get the HDC based contrastive loss, i.e.,
	\begin{align}
	\mathcal{L}_{k} = & \sum_{(i,j) \in \mathcal{P}_{k}} \mathcal{D}({f}_{i,k}^{+}, {f}_{j,k}^{+}) + \notag\\
	&\sum_{(i,j) \in \mathcal{N}_{k}} \max \{ 0, \mathcal{M} - \mathcal{D}({f}_{i,k}^{-}, {f}_{j,k}^{-})\}
	\end{align}
	
	\subsection{Definition of Hard Example}\label{sec:hard_def}
	\vspace{-0.2cm}
	
	Given the defined loss function, we follow conventional hard example mining to define the samples of large loss values as hard examples except that multiple losses will be used to mine hard examples for each sample. Because the loss distributions are different for different models and keep changing during training, it is difficult to predefine thresholds for each model when mining hard samples. Instead, we simply rank losses of all positive pairs in a mini-batch in descending order and take top $h^k$ percent samples in the ranking list as hard positive set for the model $k$. Similar strategies are adopted for hard negative example mining. Then the selected hard samples are forwarded to the later cascaded models.
	
	
	Here, we use a toy dataset with positive pairs as illustrated in Figure \ref{fig:our_work_data:pos} and negative pairs as illustrated in Figure \ref{fig:our_work_data:neg}, together with the model with \rm{K} = 3 illustrated in Figure \ref{fig:our_method} to schematically the process of hard example mining.
	\textbf{Cascade Model-1} will forward all pairs in $\mathcal{P}_0$  and $\mathcal{N}_0$, and try to push all positive points towards the anchor point while pushing all negative points away from the anchor point, and form $\mathcal{P}_1$, $\mathcal{N}_1$ (points in the $2^{nd}$ and $3^{rd}$ tier) by selecting hard samples according to its loss. Similarly, $\mathcal{P}_2$  and $\mathcal{N}_2$ (points in the $3^{rd}$ tier) are formed by \textbf{Cascade Model-2}.
	
	From the illustrated model in Figure~\ref{fig:our_method}, ensembling models in a cascaded manner brings an additional advantage of computational efficiency, since lots of computations are shared during forward pass which is efficient for both training and testing.

	\subsection{ Implementation Details }
	\vspace{-0.2cm}
	We use mini-batch SGD to optimize the loss function \eqref{cascade_sum_loss}, and adopt multi-batch~\cite{tadmor2016learning} to use all the possible pairs in a mini-batch for stable estimation of the gradient.
	Algorithm \ref{alg:cascade_framework} details the framework of our implementation for the HDC. Specifically, sampling strategy from~\cite{stanford_CVPR2016} is to construct a mini-batch of images as input, e.g., a mini-batch of 100 images are randomly sampled evenly from 10 different categories. To leverage more training samples, we further take multi-batch method~\cite{tadmor2016learning} to construct all image pairs in a mini-batch to calculate the training loss, e.g., $100^2-100 = 9900$ pairs are constructed for 100 images. With the cascaded models, an image is represented by concatenating features from all models.
	\begin{algorithm}[htb]
		\caption{Hard-Aware Deeply Cascaded Embedding.}
		\label{alg:cascade_framework}
		\begin{algorithmic}[1]
			\STATE Given training images set $\{I_i\}_{i=1}^{N}$.
			\FOR{$t=1$; $t<T$; $t++$}
			\STATE Sample a mini-batch of training images, following the method in~\cite{stanford_CVPR2016} and initialize the $\mathcal{P}_{0}$ and $\mathcal{N}_{0}$ within the mini-batch following the method in~\cite{tadmor2016learning}.
			\FOR{$k=1$; $k \le \rm{K}$; $k++$}
			\STATE Forward all the images in set $\mathcal{P}_{k-1}$ and $\mathcal{N}_{k-1}$ to $k^{th}$ model to compute the features according to Eq.\eqref{branch_feature_1} and Eq.\eqref{branch_feature_3}.
			\STATE Compute the losses for the all pairs constructed in the mini-batch according to Eq.\eqref{loss_pos_pair} and Eq.\eqref{loss_neg_pair}.
			\STATE Get the $\mathcal{P}_{k}$ and $\mathcal{N}_{k}$ by choosing the hard pairs following the method described in Section~\ref{sec:hard_def}.
			\STATE Backward and update the gradients according to corresponding parts in Eq.\eqref{cascade_bp_G} and Eq.\eqref{cascade_bp_F} for all the pairs in $\mathcal{P}_{k}$ and $\mathcal{N}_{k}$.
			\ENDFOR
			\ENDFOR
		\end{algorithmic}
	\end{algorithm}
	
	\begin{figure*}[htb]
		\centering
		\subfigure[LDA score = 0.54]{
			\addtocounter{subfigure}{0}
			\label{fig:hist_car_compare:a}
			{\includegraphics[scale=0.2]{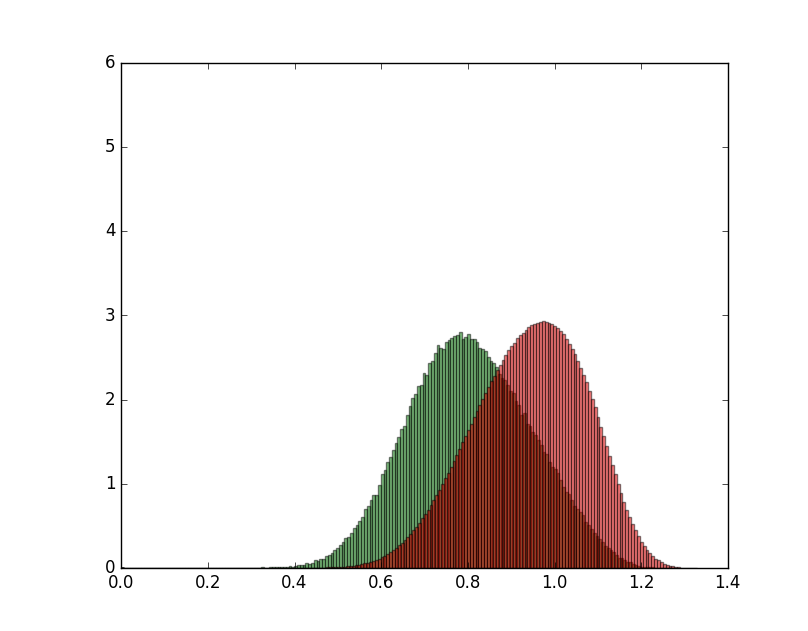}}}
		\subfigure[LDA score = 0.91]{
			\label{fig:hist_car_compare:b}
			{\includegraphics[scale=0.2]{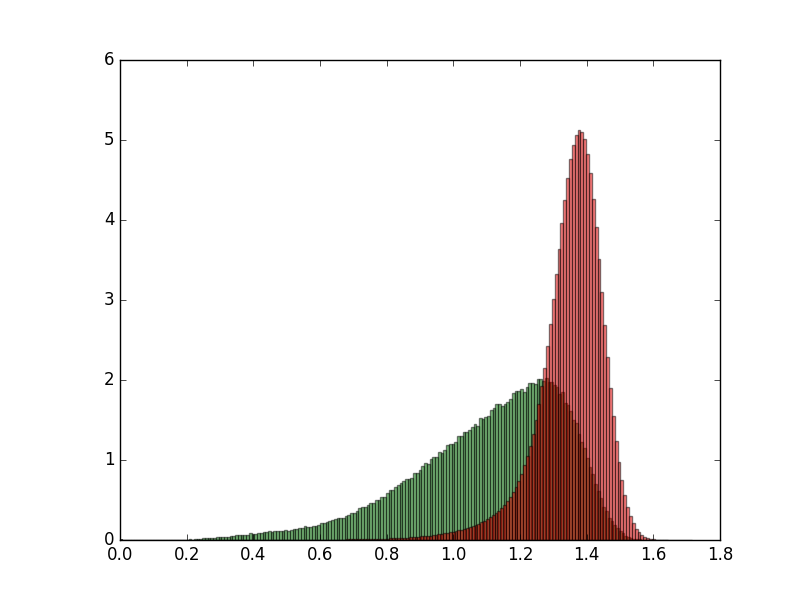}}}
		\subfigure[LDA score = 1.99]{
			\label{fig:hist_car_compare:c}
			{\includegraphics[scale=0.2]{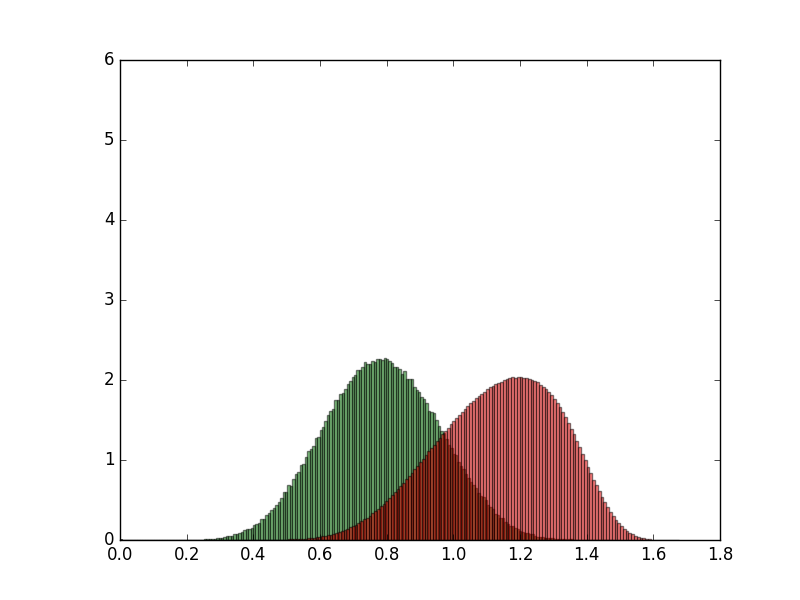}}}
		\subfigure[LDA score = 2.50]{
			\label{fig:hist_car_compare:d}
			{\includegraphics[scale=0.2]{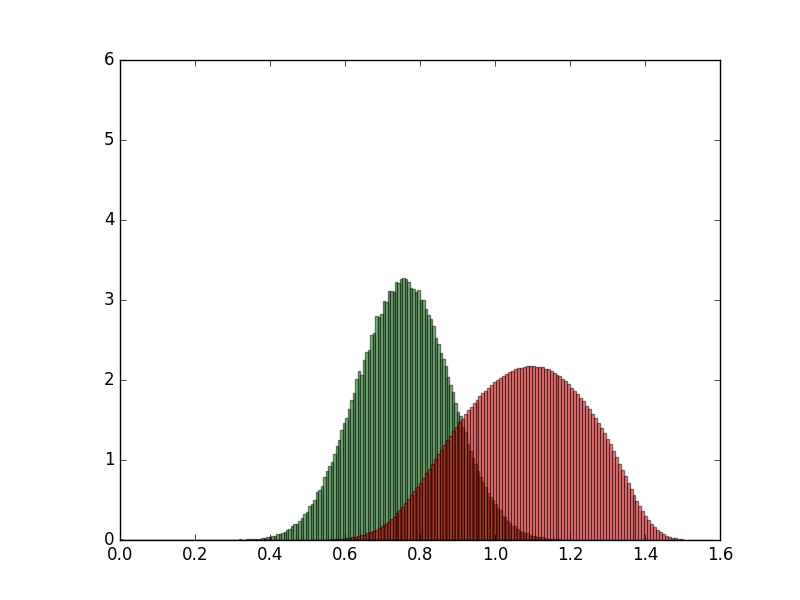}}}
		\caption{\textbf{Histograms for positive and negative distance distribution on the CARS196 test set:}
			(a) GoogLeNet/pool5$^{1024}$  (b) Contrastive$^{\dag}$$^{128}$  (c) Hard + Contrastive$^{\dag}$$^{128}$ (d) HDC + Contrastive$^{\dag}$$^{384}$. We can see that the overlap area between the 2 distributions decreases from left to right. You can check the LDA score of these methods on table \ref{table:hist_table_car} increases from left to right. }
		\label{fig:hist_car_compare}
	\end{figure*}
	
	\begin{table*}[!htb]
		\centering
		\caption{\textbf{Comparisons of the Statistics of Histograms and Recall@$K$ on CARS196 test set}. The mean and variance under the column named Positive Pairs correspond to $m^{+}$ and $v^{+}$. The mean and variance under the column named Negative Pairs correspond to $m^{-}$  and $v^{-}$.}
		\footnotesize
		\begin{tabular}{lcccccc|cccc|c}
			\hline
			& \multicolumn{6}{c|}{Recall@$K$(\%)} &\multicolumn{2}{c}{Positive Pairs } & \multicolumn{2}{c|}{Negative Pairs} \\
			\cline{2-6} \cline{7-11}
			&1 & 2 & 4 & 8 & 16 & 32 & mean & variance & mean & variance &  LDA score\\ \hline
			GoogLeNet/pool5$^{1024}$ & 40.5 & 53.0 & 65.0 & 76.3 & 86.0 & 93.1 & 0.804 & 0.019  & 0.941 & 0.016  & 0.54 \\
			Contrastive$^{\dag}$$^{128}$ & 56.0 & 67.6 & 77.0 & 84.8 & 90.5 & 94.5  & 1.110 & 0.052 & 1.350 & 0.011 & 0.91\\
			Hard + Contrastive$^{\dag}$$^{128}$  & 67.6 & 77.9 & 85.6 & 91.2 & 95.0 & 97.3 & 0.786 & 0.029 & 1.140 & 0.034 & 1.99 \\ \hline
			HDC + Contrastive-1$^{\dag}$$^{128}$  & 41.9 & 55.5 & 67.6 & 78.3 & 86.9 & 93.2 & 0.741 & 0.045 & 1.200 & 0.074  & 1.77\\
			HDC + Contrastive-2$^{\dag}$$^{128}$  & 58.0 & 70.4 & 80.2 & 87.5 & 92.9 & 96.1 & 0.660 & 0.023 & 1.050 & 0.046 & 2.20\\
			HDC + Contrastive-3$^{\dag}$$^{128}$  & 71.4 & 81.8 & 88.5 & 93.4 & \textbf{96.6} & 98.2 & 0.792 & 0.014 & 1.070 & 0.020 & 2.27\\
			HDC + Contrastive$^{\dag}$$^{384}$ & \textbf{73.7} & \textbf{83.2} & \textbf{89.5} & \textbf{93.8} & 96.5 & \textbf{98.4} & 0.756 & 0.015 & 1.080 & 0.027 & \textbf{2.50} \\ \hline
		\end{tabular}
		\label{table:hist_table_car}
	\end{table*}
	
	\begin{table*}[!htb]
		\centering
		\caption{\textbf{Comparisons of the Statistics of Histograms and Recall@$K$ on CUB-200-2011 test set}.}
		\footnotesize
		\begin{tabular}{lcccccc|cccc|c}
			\hline
			&  \multicolumn{6}{c|}{Recall@$K$(\%)} &\multicolumn{2}{c}{Positive Pairs } & \multicolumn{2}{c|}{Negative Pairs} \\
			\cline{2-6} \cline{7-11}
			& 1 & 2 & 4 & 8 & 16 & 32& mean & variance & mean & variance & LDA score\\ \hline
			HDC + Contrastive-1$^{\dag}$$^{128}_{\Box}$  & 43.4 & 55.8 & 69.1 & 80.4 & 88.1 & 93.9& 0.709 & 0.023 & 1.000 & 0.026 & 1.73\\
			HDC + Contrastive-2$^{\dag}$$^{128}_{\Box}$  & 51.9 & 63.8 & 75.1 & 84.3 & 91.2 & 95.3 & 0.637 & 0.016 & 0.919 & 0.021 & 2.15\\
			HDC + Contrastive-3$^{\dag}$$^{128}_{\Box}$  & 58.5 & 71.1 & 80.8 & 88.5 & 93.5 & 96.5 & 0.770 & 0.012 & 1.000 & 0.013 & 2.12\\
			HDC + Contrastive$^{\dag}$$^{384}_{\Box}$ & \textbf{60.7} & \textbf{72.4} & \textbf{81.9} & \textbf{89.2} & \textbf{93.7} & \textbf{96.8} & 0.741 & 0.012 & 0.989 & 0.014  & \textbf{2.37} \\ \hline
		\end{tabular}
		\label{table:hist_table_cub}
	\end{table*}

	\section{Experimental Evaluation}
	\vspace{-0.2cm}
	The proposed HDC is verified on image retrieval tasks and evaluated by two standard performance metrics, i.e., MAP and Recall@$K$. MAP~\cite{vehicleID} is the mean of average precision scores for all query images over the all the returned images. Recall@$K$ is the average recall scores over all the query images in testing set following the definition in ~\cite{song2015deep}. Specifically, for each query image, top $K$ nearest images will be returned based on some algorithm, the recall score will be 1 if at least one positive image appears in the returned $K$ images and 0 otherwise.
	
	\subsection{Datasets}
	\vspace{-0.2cm}
	Five datasets that are commonly chosen in deep metric embedding are used in our experiments. For fair comparison with the existing methods, we follow the standard protocol of train/test split.
	\begin{itemize}
		\vspace{-0.2cm}
		\item
		\emph{CARS196} dataset~\cite{stanford_cars}, which has 196 classes of cars with 16,185 images, where the first 98 classes are for training (8,054 images) and the other 98 classes are for testing (8,131 images). Both query set and database set are the test set.
		\vspace{-0.2cm}
		\item
		\emph{CUB-200-2011} dataset~\cite{wah2011caltech}, which has 200 species of birds with 11,788 images, where the first 100 classes are for training (5,864 images) and the rest of classes are for testing (5,924 images). Both query set and database set are the test set.
		\vspace{-0.2cm}
		\item
		\emph{Stanford Online Products} dataset~\cite{stanford_CVPR2016}, which has 22,634 classes with 120,053 products images, where 11,318 classes are for training (59,551 images) and 11,316 classes are for testing (60,502 images). Both query set and database set are the test set.
		\vspace{-0.2cm}
		\item
		\emph{In-shop Clothes Retrieval} dataset~\cite{DeepFashion}, which contains 11,735 classes of clothing items with 54,642 images. Following the settings in ~\cite{DeepFashion}, only 7,982 classes of clothing items with 52,712 images are used for training and testing. 3,997 classes are for training (25,882 images) and 3,985 classes are for testing (28,760 images). The test set are partitioned to query set and database set, where query set contains 14,218 images of 3,985 classes and database set contains 12,612 images of 3,985 classes.
		\vspace{-0.2cm}
		\item
		\emph{VehicleID} dataset~\cite{vehicleID} is a large-scale vehicle dataset that contains 221,763 images of 26,267 vehicles, where the training set contains 110,178 images of 13,134 vehicles and the testing set contains 111,585 images of 13,133 vehicles. Following the settings in ~\cite{vehicleID}, we use 3 test splits of different sizes constructed from the testing set. The small test set contains 7,332 images of 800 vehicles. The medium test set contains 12,995 images of 1,600 vehicles. The large test set contains 20,038 images of 2,400 vehicles.
	\end{itemize}
	
	\begin{table*}[htb]
		\centering
		\caption{\textbf{Recall@$K$(\%) on CARS196 and CUB-200-2011}.}
		\footnotesize
		\begin{tabular}{lcccccc|cccccc} \hline
			&\multicolumn{6}{c|}{CARS196} & \multicolumn{6}{c}{CUB-200-2011} \\ \cline{2-13}
			$K$ & 1 & 2 & 4 & 8 & 16 & 32 & 1 & 2 & 4 & 8 & 16 & 32  \\
			\hline
			Contrastive$^{128}$ ~\cite{bell2015learning}               & 21.7 & 32.3 & 46.1 & 58.9 & 72.2 & 83.4 & 26.4 & 37.7 & 49.8 & 62.3 & 76.4 & 85.3 \\
			Triplet$^{128}$ ~\cite{schroff2015facenet,wang2014learning} & 39.1 & 50.4 & 63.3 & 74.5 & 84.1 & 89.8 & 36.1 & 48.6 & 59.3 & 70.0 & 80.2 & 88.4\\
			LiftedStruct$^{128}$ ~\cite{song2015deep}                   & 49.0 & 60.3 & 72.1 & 81.5 & 89.2 & 92.8  & 47.2 & 58.9 & 70.2 & 80.2 & 89.3 & 93.2\\
			\hline
			Binomial Deviance$^{512}$~\cite{UstinovaNIPS16}   & - & - & - & - & - & -  & 52.8 & 64.4 & 74.7 & 83.9 & 90.4 & 94.3 \\
			Histogram Loss$^{512}$~\cite{UstinovaNIPS16}       & - & - & - & - & - & -  & 50.3 & 61.9 & 72.6 & 82.4 & 88.8 & 93.7\\
			\hline
			HDC + Contrastive$^{\dag}$$^{384}$  & \textbf{73.7} & \textbf{83.2} & \textbf{89.5} & \textbf{93.8} & \textbf{96.7} & \textbf{98.4} & \textbf{53.6} & \textbf{65.7} & \textbf{77.0} & \textbf{85.6} & \textbf{91.5} & \textbf{95.5} \\
			\hline
			\hline
			PDDM + Triplet$_{\Box}^{128}$ ~\cite{cuhk_nips2016}        & 46.4 & 58.2 & 70.3 & 80.1 & 88.6 & 92.6  & 50.9 & 62.1 & 73.2 & 82.5 & 91.1 & 94.4\\
			PDDM + Quadruplet$_{\Box}^{128}$ ~\cite{cuhk_nips2016}      & 57.4 & 68.6 & 80.1 & 89.4 & 92.3 & 94.9 & 58.3 & 69.2 & 79.0 & 88.4 & 93.1 & 95.7\\
			\hline
			HDC + Contrastive$^{\dag}$$_{\Box}^{384}$ & \textbf{83.8} & \textbf{89.8} & \textbf{93.6} & \textbf{96.2} & \textbf{97.8} & \textbf{98.9} & \textbf{60.7} & \textbf{72.4} & \textbf{81.9} & \textbf{89.2} & \textbf{93.7} & \textbf{96.8} \\
			\hline
			\hline
			Npairs$_{\boxplus}$~\cite{sohn2016improved}   & 71.1 & 79.7 & 86.5 & 91.6 & - & -  & 50.9 & 63.3 & 74.3 & 83.2 & - & -\\
			\hline
			HDC + Contrastive$^{\dag}$$_{\boxplus}^{384}$ & \textbf{75.0} & \textbf{83.9} & \textbf{90.3} & \textbf{94.3} & \textbf{96.8} & \textbf{98.4} & \textbf{54.6} & \textbf{66.8} & \textbf{77.6} & \textbf{85.9} & \textbf{91.7} & \textbf{95.6} \\
			\hline
		\end{tabular}
		\label{table:recall_car_cub}
	\end{table*}
	
	\subsection{Experiment Setup}
	\vspace{-0.2cm}
	We choose GoogLeNet~\cite{Szegedy2015Going} as our model for retrieval tasks. Since GoogLeNet has three output classifiers(two auxiliary classifiers from intermediate layers), HDC adopts them as three cascaded sub-networks corresponding to the three rows illustrated in Figure 2. i.e., G1 contains the layers from the input to "Inception(4a)" inclusively before the first classifier. We initialize the weights from the network pretrained on ImageNet ILSVRC-2012~\cite{russakovsky2015imagenet}. We use the same hyper parameters in all experiments without specifically tuning. Specifically, K is set to 3, {$\lambda_1$=$\lambda_2$=$\lambda_3$}=1, $\{h^{1}, h^{2}, h^{3}\}$ = $\{100,50,20\}$, mini-batch size is 100, margin parameter $\mathcal{M}$ is set to 1, the initial learning rate starts from 0.01 and is divided by 10 every 3-5 epoches, and we train models for at most 15 epoches. The other settings follow the same protocol in ~\cite{stanford_CVPR2016}. The embedding dimensions of all the cascade models in our HDC are 128, so the embedding dimension of the ensembled model is 384. The code is publicly available at {\footnotesize{\url{https://github.com/PkuRainBow/Hard-Aware-Deeply-Cascaded-Embedding_release}}}\par
	
	%
	\subsection{Comparison with Baseline}
	\vspace{-0.2cm}
	We name different methods with superscript and subscript to denote their specific settings, the number in superscript denotes the dimension used by the method, the subscript $\Box$ denotes bounding boxes are used during training and testing. Different from the original Contrastive$^{128}$~\cite{bell2015learning}, we use the Contrastive$^{\dag}$$^{128}$ to denote the contrastive loss computed with multi-batch~\cite{tadmor2016learning} method. \par
	
	To directly verify the effectiveness of HDC, we first design several baseline methods including: (1) \textbf{GoogLeNet/pool5$^{1024}$} uses the feature vector directly outputted from pool5 of the pre-trained GoogLeNet, (2) \textbf{Contrastive$^{\dag}$$^{128}$} uses contrastive loss without hard example mining, (3) \textbf{Hard + Contrastive$^{\dag}$$^{128}$} combines the contrastive loss and hard example mining. In addition to report our method named as \textbf{HDC + Contrastive$^{\dag}$$^{384}$}, we also report the performance of sub models learned in our method, i.e., \textbf{HDC + Contrastive-1$^{\dag}$$^{128}$}, \textbf{HDC + Contrastive-2$^{\dag}$$^{128}$} and \textbf{HDC + Contrastive-3$^{\dag}$$^{128}$}. Hard+Contrastive uses the same network architecture as HDC+Contrastive-3, i.e., \{G1,G2,G3,F3\}. Only top 50 percent examples with larger loss are chosen as hard examples to update the model. The results of these methods on CARS196 are summarized in Table ~\ref{table:hist_table_car}. Obviously, training on the target dataset brings significant performance improvement comparing with \textbf{GoogLeNet/pool5$^{1024}$}, hard example mining further brings more performance gain, while the hard aware method achieves the best performance. \textbf{HDC + Contrastive-3$^{\dag}$$^{128}$} is much better than the traditional \textbf{Hard + Contrastive$^{\dag}$$^{128}$} as the shallow modules of the model are also trained by hard samples mined by shallow models. Our method in the last row achieves the best result comparing with all baselines, which verifies the effectiveness of the hard-aware sample mining.
	
	Figure \ref{fig:hist_car_compare} shows the distance distributions of positive pairs and negative pairs following~\cite{UstinovaNIPS16}, where green area represents the distance distributions of positive pairs while red area for negative pairs. Our method has the smallest overlapping area, and better separates positive pairs and negative pairs in the embedding space. We also calculate the LDA score which measures the distance between two distributions to quantitatively compare the difference, i.e.,
	\begin{align}
	score &= \frac{\vert m^{-}-m^{+} \vert ^{2}}{ v^{+} +  v^{-} }
	\end{align}
	where $m^{+}$ and $m^{-}$ are the mean distance of positive pairs and negative pairs, $v^{+}$ and $v^{-}$ are the variance of the distances of positive pairs and negative pairs. The results  on CARS196 are reported in the right part of Table \ref{table:hist_table_car}. It can be observed that the retrieval performance measured by Recall@$K$ positively correlates with LDA score, and our method achieves the highest LDA score 2.50. We also conduct experiment on CUB-200-2011 and report the results in Table \ref{table:hist_table_cub}, where the conclusion is the same on CARS196.
	
	\subsection{Comparison with state-of-the-art}
	\vspace{-0.2cm}
	We compare our method with state-of-the-art methods on the five datasets.
	On the CARS196, CUB-200-2011 and Stanford Online Products datasets: (1) \textbf{LiftedStruct$^{128}$}~\cite{song2015deep} uses a novel structured prediction objective on the lifted dense pairwise distance matrix. (2) \textbf{PDDM + Triplet$_{\Box}^{128}$} ~\cite{cuhk_nips2016} combines Position-Dependent Deep Metric units (PDDM) and Triplet Loss. (3) \textbf{PDDM + Quadruplet$_{\Box}^{128}$} ~\cite{cuhk_nips2016} combines the PDDM with Quadruplet Losses proposed in~\cite{Law_2013_ICCV}. (4) \textbf{Histogram Loss$^{512}$}~\cite{UstinovaNIPS16} is penalizing the overlap between distributions of distances of positive pairs and negative pairs.(5) \textbf{Binomial Deviance$^{512}$}~\cite{UstinovaNIPS16} is used to evaluate the cost between similarities and labels, which is proved robust to outliers. (6)\textbf{Npairs$_{\boxplus}$}~\cite{sohn2016improved} uses multi-class N-pair loss by generalizing triplet loss by allowing joint comparison among more than one negative examples. The subscript ${\boxplus}$ means using multiple crops when testing, while all the other methods use single crop except \textbf{Npairs$_{\boxplus}$}. All these methods use GoogLeNet as the base model, which is the same as our method. 
	\par
	
	\begin{table}[!htb]
		\centering
		\caption{\textbf{Recall@$K$(\%) on In-shop Clothes Retrieval Dataset}.}
		\resizebox{\linewidth}{!}{
			\begin{tabular}{lcccccc}
				\hline
				$K$ & 1 & 10 & 20 & 30 & 40 & 50 \\ \hline
				FashionNet + Joints$^{4096}$~\cite{DeepFashion} & 41.0 & 64.0 & 68.0 & 71.0 & 73.0 & 73.5\\
				FashionNet + Poselets$^{4096}$~\cite{DeepFashion} & 42.0 & 65.0 & 70.0 & 72.0 & 72.0 & 75.0\\
				FashionNet$^{4096}$ ~\cite{DeepFashion} & 53.0 & 73.0 & 76.0 & 77.0 & 79.0 & 80.0\\\hline
				HDC + Contrastive$^{\dag}$$^{384}$ & \textbf{62.1} & \textbf{84.9} & \textbf{89.0} & \textbf{91.2} & \textbf{92.3} & \textbf{93.1}\\
				\hline
			\end{tabular}
		}
		\label{table:deep_fashion_inshop}
	\end{table}

	On VehicleID and In-shop Clothes Retrieval datasets : (1) \textbf{CCL + Mixed Diff$^{1024}$}~\cite{vehicleID} uses the Coupled Cluster Loss and Mixed Difference Network Structure. (2) \textbf{FashionNet}~\cite{DeepFashion} simultaneously learns the landmarks and attributes of the images using \textit{VGG-16} ~\cite{simonyan2014very}. To test the generality of our method, we use the same hyper-parameters without specifically tuning on these datasets.
	
	\begin{figure*}[htb]
		\centering
		\subfigure[]{
			\label{fig:result:products}
			{\includegraphics[scale=0.4]{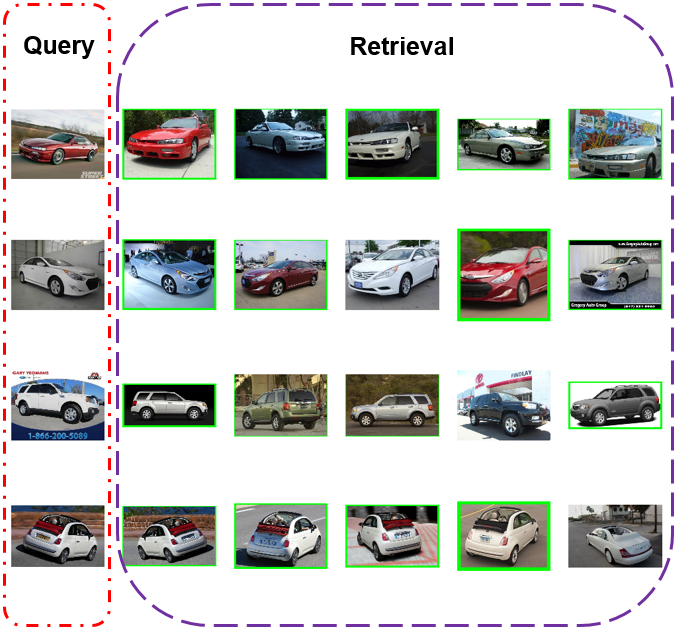}}}
		\hspace{15mm}
		\subfigure[]{
			\label{fig:result:cars}
			{\includegraphics[scale=0.4]{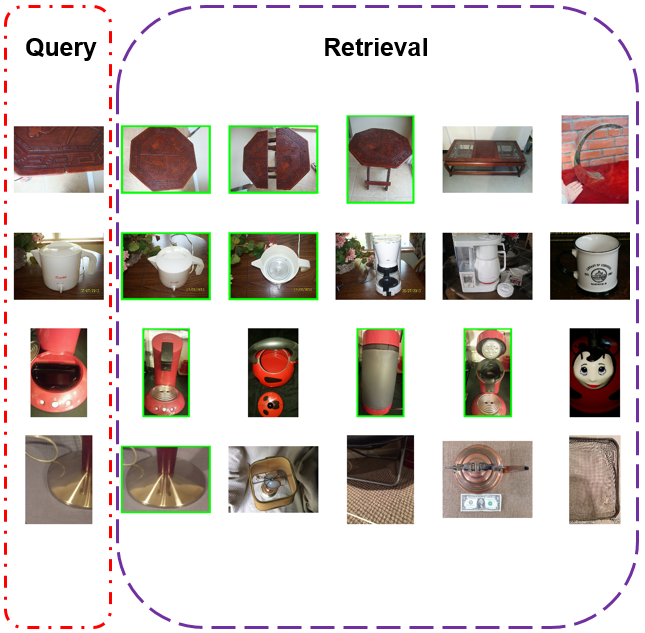}}}
		\label{fig:result}
		\caption{\textbf{Retrieval Results on CARS196 and Stanford Online Products:} (a) CARS196. (b) Stanford Online Products.}
		\vspace{-0.35cm}
	\end{figure*}
	
	Table \ref{table:recall_car_cub} quantifies the advantages of our method on both CARS196 and CUB-200-2011. We conduct three groups of experiments to ensure fairness as different methods adopt different settings, i.e., with/without bounding boxes and with multiple crops when testing. \textbf{PDDM + Triplet$_{\Box}^{128}$} and \textbf{PDDM + Quadruplet$_{\Box}^{128}$} both use the images cropped with the annotated bounding boxes as training set and test set. With bounding boxes, cluttered backgrounds are removed and better performance is expected. HDC + Contrastive$^{\dag}$$_{\Box}^{384}$ shows significant performance gain both on CARS196 and CUB-200-2011. On CARS196, we improve the Recall@1 score from 57.4\% to 83.8\%. CUB-200-2011 is more challenging than CARS196 as the car is rigid while birds have more variations. We get 2.4\% absolute improvement on CUB-200-2011. \textbf{Histogram Loss$^{512}$} and \textbf{Binomial Deviance$^{512}$} are trained without bounding boxes, for fair comparison, we also validate our method without using bounding boxes. HDC + Contrastive$^{\dag}$$^{384}$ outperforms all methods without using bounding boxes on both datasets, and has even better results than methods using bounding boexes on CARS196. Besides, we test our method when using multiple crops for test. HDC + Contrastive$^{\dag}$$_{\boxplus}^{384}$ also achieves state-of-the-art performance compared with \textbf{Npairs$_{\boxplus}$}. \par

	\begin{table}[htb]
		\centering
		\caption{\textbf{Recall@$K$(\%) on Stanford Online Products}.}
		\footnotesize
		\begin{tabular}{lcccc} \hline
			$K$ & 1 & 10 & 100 & 1000  \\
			\hline
			Contrastive$^{128}$~\cite{bell2015learning}                 & 42.0 & 58.2 & 73.8 & 89.1\\
			Triplet$^{128}$~\cite{schroff2015facenet,wang2014learning}  & 42.1 & 63.5 & 82.5 & 94.8\\
			LiftedStruct$^{128}$~\cite{song2015deep}                    & 60.8 & 79.2 & 91.0 & 97.3\\
			LiftedStruct$^{512}$~\cite{song2015deep}            & 62.1 & 79.8 & 91.3 & 97.4\\
			\hline
			Binomial Deviance$^{512}$~\cite{UstinovaNIPS16}     & 65.5 & 82.3 & 92.3 & 97.6\\
			Histogram Loss$^{512}$~\cite{UstinovaNIPS16}        & 63.9 & 81.7 & 92.2 & 97.7\\
			\hline
			HDC + Contrastive$^{\dag}$$^{384}$  & \textbf{69.5} & \textbf{84.4} & \textbf{92.8} & 97.7\\
			\hline
			\hline
			Npairs$_{\boxplus}$~\cite{sohn2016improved}       & 67.7 & 83.8 & 92.9 & 97.8\\
			\hline
			HDC + Contrastive$^{\dag}$$_{\boxplus}^{384}$ & \textbf{70.1} & \textbf{84.9} & \textbf{93.2} & 97.8\\
			\hline
		\end{tabular}
		\label{table:recall_products}
	\end{table}
	
	Table \ref{table:recall_products} reports the results on Stanford Online Products. Stanford Online Products suffers the problem of large number of categories and few images per category, which is very different from CARS196 and CUB-200-2011. Our method achieves 4\% absolute improvements over previous state-of-the-art methods measured by Recall@1. When testing with multiple crops, HDC + Contrastive$^{\dag}$$_{\boxplus}^{384}$ further improves the Recall@1 from 67.7\% to 70.1\%. Figure \ref{fig:result:products} shows some retrieval results on Stanford Online Products with features learned by HDC + Contrastive$^{\dag}$$^{384}$.
	
	Similar to the Stanford Online Products, DeepFashion In-shop Clothes and VehicleID also suffer the problem of limited images in each class and large number of classes. Table~\ref{table:deep_fashion_inshop} and~\ref{table:map_vehicle_id} compare the results on the two datasets, where our method outperforms state-of-the-art methods by a large margin.
	
	Through extensive empirical comparisons on various datasets under different settings, we show that our method is general and can achieve better performance.
	
	\begin{table}[htb]
		\centering
		\caption{\textbf{MAP of Vehicle Retrieval Task}.}
		\footnotesize
		\begin{tabular}{lccc}
			\hline
			MAP & Small & Medium & Large \\ \hline
			VGG + Triplet Loss$^{1024}$ & 0.444 & 0.391 & 0.373 \\
			VGG + CCL$^{1024}$ (~\cite{vehicleID}) & 0.492 & 0.448 & 0.386\\
			Mixed Diff + CCL$^{1024}$ (~\cite{vehicleID}) & 0.546 & 0.481 & 0.455\\\hline
			GoogLeNet/pool5$^{1024}$ & 0.418 & 0.392 & 0.347\\
			HDC + Contrastive$^{\dag}$$^{384}$ & \textbf{0.655} & \textbf{0.631} & \textbf{0.575}\\ \hline
		\end{tabular}
		\label{table:map_vehicle_id}
	\end{table}
	
	\section{Conclusions}
	\vspace{-0.2cm}
	In this paper, we propose a novel Hard-Aware Deeply Cascaded Embedding to consider both hard levels of samples and the complexities of models. Different from training three separated models, our design ensembles a set of models with increasing complexities in a cascaded manner and shares most of the computation among models. Samples with different hard levels are mined accordingly using the models with adequate complexities. Controlled experimental results demonstrate the advantages by the hard-aware design, and extensive comparisons on five benchmarks further verify the effectiveness of the proposed method in learning deep metric embedding.
	\par
	Currently, the method is verified by three cascaded models with increasing complexities, in the future, we would further improve the method by cascading more models and increasing complexities in a smoother way. And we would also try to combine our method with other loss functions in the future work.

	\section*{Acknowledgements}
	\vspace{-0.2cm}
	This work is partially supported by the National Key Basic Research Project of China (973 Program)
	under Grant 2015CB352303 and the National Nature Science Foundation of China under Grant 61671027.

	{\small
		\bibliographystyle{ieee}
		\bibliography{egbib}
	}
	
\end{document}